\documentclass[11pt,a4paper]{article}
\usepackage[hyperref]{naaclhlt2019}
\usepackage{times}
\usepackage{latexsym}
\usepackage{xspace}
\usepackage{subcaption,siunitx,booktabs}
\usepackage{multirow}
\usepackage{makecell}
\usepackage{graphicx}
\usepackage{mathtools}
\usepackage{amssymb}
\usepackage{tcolorbox}
\usepackage{pagecolor}
\usepackage{amsmath}
\usepackage{enumitem}
\usepackage{makecell}
\usepackage{xcolor}
\usepackage{pifont}
\newcommand{\cmark}{\ding{51}}
\newcommand{\xmark}{\ding{55}}
\usepackage{footnote}
\usepackage{url}

\tcbuselibrary{skins}
\tcbset{colframe=white}

\aclfinalcopy 

\title{Using Natural Language Relations between Answer Choices\\ for Machine Comprehension}

\author{  Rajkumar Pujari\\
  Department of Computer Science \\
  Purdue University \\
  {\tt rpujari@purdue.edu} \\\And
    Dan Goldwasser\\
  Department of Computer Science \\
  Purdue University \\
  {\tt dgoldwas@purdue.edu} \\}

\date{}

\begin{document}
\maketitle

\begin{abstract}
When evaluating an answer choice for Reading Comprehension task, other answer choices available for the question and the answers of related questions about the same paragraph often provide valuable information. In this paper, we propose a method to leverage the natural language relations between the answer choices, such as entailment and contradiction, to improve the performance of machine comprehension. We use a stand-alone question answering (QA) system to perform QA task and a Natural Language Inference (NLI) system to identify the relations between the choice pairs. Then we perform inference using an Integer Linear Programming (ILP)-based relational framework to re-evaluate the decisions made by the standalone QA system in light of the relations identified by the NLI system. We also propose a multitask learning model that learns both the tasks jointly.   
\end{abstract}

\section{Introduction}\label{sec:intro}
\indent Given an input text and a set of related questions with multiple answer choices, the reading comprehension (RC) task evaluates the correctness of each answer choice. Current approaches to the RC task quantify the relationship between each question and answer choice independently and pick the highest scoring option. In this paper, we follow the observation that when humans approach such RC tasks, they tend to take a holistic view ensuring that their answers are consistent across the given questions and answer choices. In this work we attempt to model these pragmatic inferences, by leveraging the \textit{entailment} and \textit{contradiction} relations between the answer choices to improve machine comprehension. To help clarify these concepts, consider the following examples:
\fbox{\parbox{\dimexpr\linewidth-2\fboxsep-2\fboxrule\relax}{
\textit{How can the military benefit from the existence of the CIA?}\\
$c_1$: They can use them\\
$c_2$: \textcolor{red}{These agencies are keenly attentive to the military's strategic and tactical requirements} (\xmark)\\
$c_3$: \textcolor{blue}{The CIA knows what intelligence the military requires and has the resources to obtain that intelligence} (\cmark) }}

The above example contains multiple correct answer choices, some are easier to capture than others. For example, identifying that $c_3$ is true might be easier than $c_2$ based on its alignment with the input text. However, capturing that $c_3$ entails $c_2$ allows us to predict $c_2$ correctly as well. Classification of the answer in red (marked \xmark) could be corrected using the blue (marked \cmark) answer choice.\\

\begin{figure}[!tbh]
    \centering
        \includegraphics[width=6.5cm]{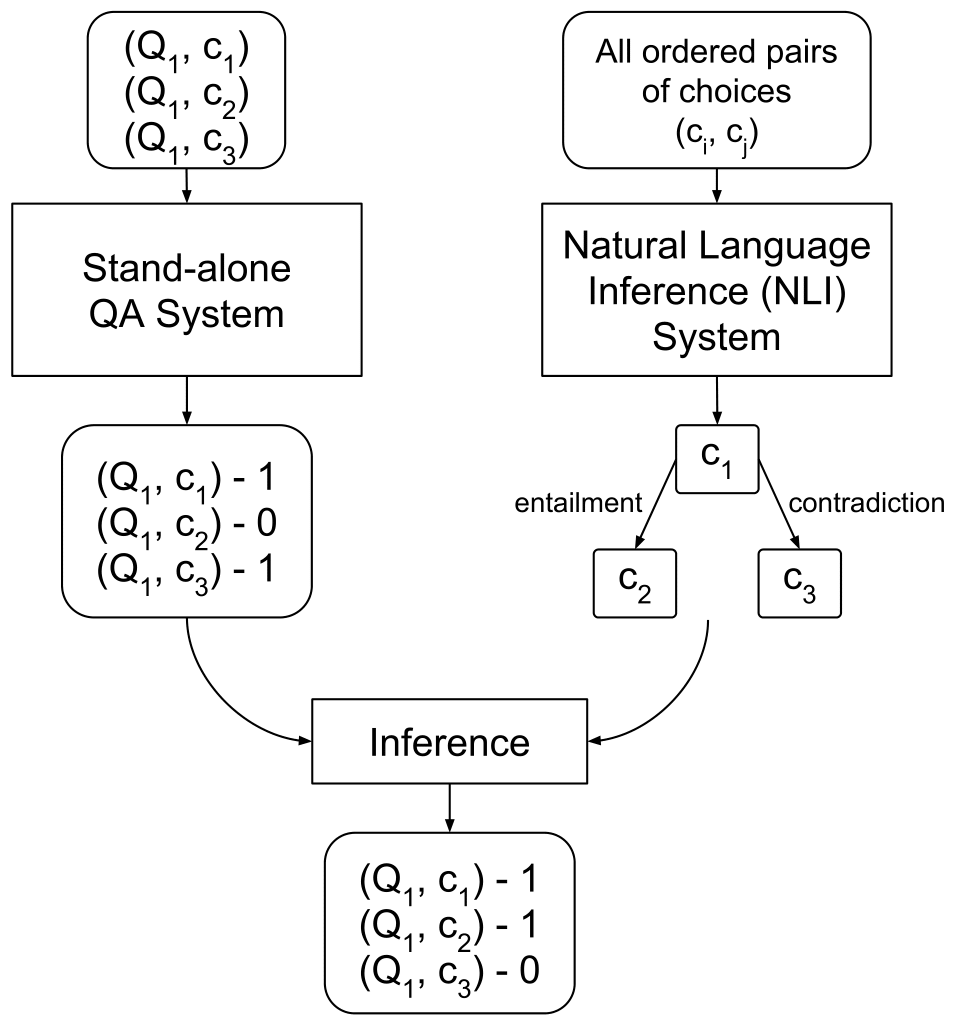}
        \caption{Proposed Approach}
        \label{fig:app}
\end{figure}

\noindent \fbox{\parbox{\dimexpr\linewidth-2\fboxsep-2\fboxrule\relax}{
Q1: \textit{When were the eggs added to the pan to make the omelette?}\\
$c_1^1$: When they turned on the stove\\
\textcolor{blue}{$c_2^1$: When the pan was the right temperature} (\cmark)\\
Q2: \textit{Why did they use stove to cook omelette?}\\
$c_1^2$: They didn't use the stove but a microwave\\
\textcolor{red}{$c_2^2$: Because they needed to heat up the pan} (\xmark) }} \\
\indent Similarly, answering Q1 correctly helps in answering Q2. Our goal is to leverage such inferences for machine comprehension.\\
\indent Our approach contains three steps. First, we use a stand-alone QA system to classify the answer choices as true/false. Then, we classify the relation between each pair of choices for a given question as \textit{entailment}, \textit{contradiction} or \textit{neutral}. Finally, we re-evaluate the labels assigned to choices using an Integer Linear Programming based inference procedure. We discuss different training protocols and representation choices for the combined decision problem. An overview is in figure \ref{fig:app}. \footnote{Code and data are available \href{https://github.com/pujari-rajkumar/naacl2019}{here}} .\\
\indent We empirically evaluate on two recent datasets, MultiRC \citep{multirc18} and SemEval-2018 task-11 \citep{semeval-t11} and show that it improves machine comprehension in both.

\section{Related Work}\label{sec:related}
Recently, several QA datasets have been proposed to test machine comprehension \citep{mctest2013,babi2015,squad2016,news-qa2016,msmacro2016}.  \citet{yatskar2018} showed that a high performance on these datasets could be achieved without necessarily achieving the capability of making commonsense inferences. \citet{parallel-hierarchical-2016}, \citet{kumar2016ask}, \citet{liu2017gated}, \citet{min-context2018} and \citet{xiong2016dynamic} proposed successful models on those datasets. To address this issue, new QA datasets which require commonsense reasoning have been proposed \cite{multirc18,semeval-t11,mihaylov2018can}. Using common sense inferences in Machine Comprehension is a far from solved problem. There have been several attempts in literature to use inferences to answer questions. Most of the previous works either attempt to infer the answer from the given text \cite{sachan2016, rc-graph2018} or an external commonsense knowledge base \cite{univ-schema2017, P18-1076, D18-1454, Weissenborn17}. 

While neural models can capture some dependencies between choices through shared representations,  to the best of our knowledge, inferences capturing the dependencies between answer choices or different questions have been not explicitly modeled. 

\section{Model}\label{sec:model}
\indent Formally, the task of machine comprehension can be defined as: given text $\mathcal{P}$ and a set of $n$ related questions $\mathcal{Q}$ = $\{q_1, q_2, \ldots{}, q_n\}$ each having $m$ choices $\mathcal{C}$ = $\{c_1^i, c_2^i,  \ldots{}, c_m^i\} \forall q_i \in \mathcal{Q}$, the task is to assign true/false value for each choice $c_j^i$. 
\subsection{Model Architecture}\label{subsec:model}
\begin{figure}[!h]
    \centering
        \includegraphics[width=5.9cm]{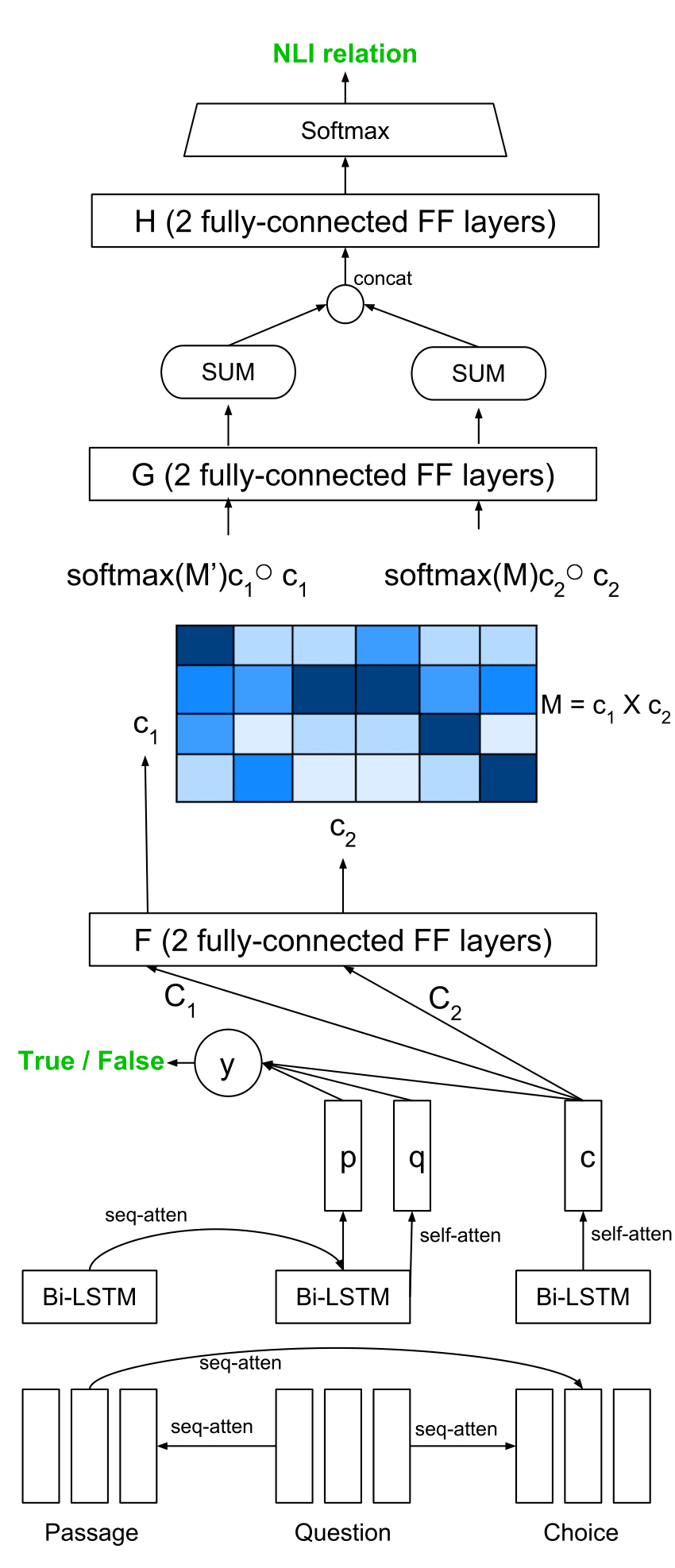}
        \caption{Architecture of the Joint Model}
        \label{fig:joint_model}
\end{figure}

\indent Our model consists of three separate systems, one for each step, namely, the stand-alone question answering (QA) system, the Natural Language Inference (NLI) system and the inference framework connecting the two. First, we assign a true/false label to each question-choice pair using the stand-alone QA system along with an associated confidence score $s_1$. Consequently, we identify the natural language relation (entailment, contradiction or neutral) between each ordered pair of choices for a given question, along with an associated confidence score $s_2$. Then, we use a relational framework to perform inference using the information obtained from the stand-alone QA and the NLI systems. Each of the components is described in detail in the following sub-sections.\\
\indent We further propose a joint model whose parameters are trained jointly on both the tasks. The joint model uses the answer choice representation generated by the stand-alone QA system as input to the NLI detection system. The architecture of our joint model is shown in figure \ref{fig:joint_model}.
\subsubsection{Stand-alone QA system}
We use the TriAN-single model proposed by \citet{yunfudao} for SemEval-2018 task-11 as our stand-alone QA system. We use the implementation\footnote{\url{https://github.com/intfloat/commonsense-rc}} provided by \citet{yunfudao} for our experiments. The system is a tri-attention model that takes passage-question-choice triplet as input and produces the probability of the choice being true as its output.
\subsection{NLI System}
Our NLI system is inspired from decomposable-attention model proposed by \citet{parikh-nli}. We modified the architecture proposed in \citet{parikh-nli} to accommodate the question-choice pairs as opposed to sentence pairs in the original model. We added an additional sequence-attention layer for the question-choice pairs to allow for the representation of both the answer choice and the question. Sequence-attention is defined in \citet{yunfudao} as:\\[-22pt]
\begin{equation}\label{eq:seq}
    \begin{aligned}
    Att_{seq}(\textbf{u}, \{\textbf{v}_i\}_{i=1}^{n}) = \sum_{i=1}^{n} \alpha_{i}\textbf{v}_i\\
    \alpha_{i} = softmax_i(f(\textbf{W}_1 \textbf{u})^T f(\textbf{W}_1 \textbf{v}_i))
    \end{aligned}
\end{equation}
where $\textbf{u}$ and $\textbf{v}_i$ are word embeddings, $\textbf{W}_1$ is the associated weight parameter and $f$ is non-linearity. Self-attention is $Att_{seq}$ of a vector onto itself.\\
\indent The embedding of each word in the answer choice is attended to by the sequence of question word embeddings. We use pre-trained GloVe \citep{pennington2014glove} embeddings to represent the words. The question-attended choices are then passed through the decomposable-attention layer proposed in \citet{parikh-nli}.
\subsubsection{Inference using DRAIL}
We use Deep Relational Learning (DRaiL) framework proposed by \citet{drail2016} to perform the final inference. The framework allows for declaration of predicate logic rules to perform relational inference. The rules are scored by the confidence scores obtained from the stand-alone QA and the NLI systems. DRaiL uses an Integer Linear Programming (ILP) based inference procedure to output binary prediction for each of the choices. We use the following constraints for our inference:
\begin{enumerate}[noitemsep]
    \item{$c_{i}$ is true \& $c_{i}$ entails $c_{j}$ $\implies$ $c_{j}$ is true.}
    \item{$c_{i}$ is true \& $c_{i}$ contradicts $c_{j}$ $\implies$ $c_{j}$ is false.}
\end{enumerate}
\indent On the MultiRC dataset, we use the dependencies between the answer choices for a given question. On SemEval dataset, we use the dependencies between different questions about the same paragraph.
\subsection{Joint Model}
\indent The design of our joint model is motivated by the two objectives: 1) to obtain a better representation for the question-choice pair for NLI detection and 2) to leverage the benefit of multitask learning. Hence, in the joint model, choice representation from stand-alone QA system is input to the decomposable-attention layer of the NLI system.\\
\indent The joint model takes two triplets ($p$, $q_i$, $c_i$) and ($p$, $q_j$, $c_j$) as input. It outputs a \texttt{true}/\texttt{false} for each choice and an NLI relation (entailment, contradiction or neutral) between the choices. The representations for passage, question and choice are obtained using Bi-LSTMs. The hidden states of the Bi-LSTM are concatenated to generate the representation. This part of the model is similar to TriAN model proposed in \citet{yunfudao}. The choice representations of $c_i$ and $c_j$ are passed as input to the decomposable attention layer proposed in \citet{parikh-nli}. The architecture of the joint model is shown in figure \ref{fig:joint_model}.
\subsection{Training} \label{sec:training}
\indent We train the stand-alone QA system using the MultiRC and SemEval datasets for respective experiments. We experiment with $2$ different training settings for the NLI system. In the first setting, we use SNLI dataset~\cite{D15-1075} to train the NLI system. The sequence-attention layer is left untrained during this phase. Hence, we only use the answer choice and do not consider the question for NLI detection.\\
\noindent \textbf{Self-Training:} Subsequently, to help the system adapt to our settings, we devise a self-training protocol over the RC datasets to train the NLI system. Self-training examples for the NLI system were obtained using the following procedure: if the SNLI-trained NLI model predicted entailment and the gold labels of the ordered choice pair were \texttt{true}-\texttt{true}, then the choice pair is labeled as \textit{entailment}. Similarly, if the SNLI-trained NLI model predicted contradiction and the gold labels of the ordered choice pair were \texttt{true}-\texttt{false}, then the choice pair is labeled as \textit{contradiction}. This is noisy labelling as the labels do not directly indicate the presence of NLI relations between the choices. The NLI model was additionally trained using this data.
\begin{table}[!h]
\begin{tabular}{|l|c|c|c|}
\hline
\textbf{\small Model}           & \textbf{\small Entailment} & \textbf{\small Contradiction}    & \textbf{\small Overall} \\ \hline
\textbf{\small $NLI_{SNLI}$}    & 40.80            & 74.25                 & 55.11            \\ \hline
\textbf{\small $NLI_{MultiRC}$} & 57.30            & 69.22                 & 66.31            \\ \hline
\end{tabular}
\caption{\small Accuracy of entailment and contradiction detection on the development set of self-training data for NLI model trained on SNLI data ($NLI_{SNLI}$) vs training set of self-training data ($NLI_{MultiRC}$)}
\label{tab:self_tr}
\end{table}

\vspace{-12pt}
\indent To train the joint model we use ordered choice pairs, labeled as \textit{entailment} if the gold labels are \texttt{true}-\texttt{true} and labeled as \textit{contradiction} if the gold labels are \texttt{true}-\texttt{false}. This data was also used to test the effectiveness of the self-training procedure. The results on the development set of MultiRC dataset are in table \ref{tab:self_tr}.\\
\indent  The NLI model trained on SNLI dataset achieves $55.11\%$ accuracy. Training the $NLI$ model on the data from MultiRC data increases the overall accuracy to $66.31\%$. Further discussion about self-training is provided in section \ref{sec:disc}.

\section{Experiments}\label{sec:expt}
We perform experiments in four phases. In the first phase, we evaluate the stand-alone QA system. In the second phase, we train the NLI system on SNLI data and evaluate the approach shown in figure \ref{fig:app}. In the third phase, we train the NLI system using the self-training data. In the fourth phase, we evaluate the proposed joint model. We evaluate all models on MultiRC dataset. The results are shown in table \ref{tab:multirc_results}. We evaluate the joint model on SemEval dataset, shown in table \ref{tab:semeval_results}.
\subsection{Datasets}\label{sec:data}
We use two datasets for our experiments, MultiRC dataset\footnote{\url{http://cogcomp.org/multirc/}} and the SemEval 2018 task 11 dataset\footnote{\url{https://competitions.codalab.org/competitions/17184}}. MultiRC dataset consisted of a training and development set with a hidden test set. We split the given training set into training and development sets and use the given development set as test set.\\
\indent Each question in the MultiRC dataset has approximately $5$ choices on average. Multiple of them may be \texttt{true} for a given question. The training split of MultiRC consisted of $433$ paragraphs and $4,853$ questions with $25,818$ answer choices. The development split has $23$ paragraphs and $275$ questions with $1,410$ answer choices. Test set has $83$ paragraphs and $953$ questions with $4,848$ answer choices.\\
\indent SemEval dataset has $2$ choices for each question, exactly one of them is \texttt{true}. The training set consists of $1,470$ paragraphs with $9,731$ questions. The development set has $219$ paragraphs with $1,411$ questions. And the test set has $430$ paragraphs with $2,797$ questions.
\subsection{Evaluation Metrics} \label{sec:eval}
\indent For MultiRC dataset, we use two metrics for evaluating our approach, namely $EM0$ and $EM1$. $EM0$ refers to the percentage of questions for which all the choices have been correctly classified. $EM1$ is the the percentage of questions for which at most one choice is wrongly classified. For the SemEval dataset, we use \textit{accuracy} metric.
\subsection{Results}\label{sec:results}
\indent Results of our experiments are summarized in tables \ref{tab:multirc_results} \& \ref{tab:semeval_results}. $EM0$ on MC task improves from $18.15\%$ to $19.41\%$ when we use the NLI model trained over SNLI data and it further improves to $21.62\%$ when we use MultiRC self-training data. Joint model achieves $20.36\%$ on $EM0$ but achieves the highest $EM1$ of $57.08\%$. Human $EM0$ is $56.56\%$. 
\begin{table}[!h]
\centering
\begin{tabular}{|l|l|l|}
\hline
\textbf{Method}                & \textbf{EM0} & \textbf{EM1} \\ \hline
\textbf{Stand-alone QA}        & 18.15        & 52.99        \\ \hline
\textbf{QA + $NLI_{SNLI}$}    & 19.41        & 56.13        \\ \hline
\textbf{QA + $NLI_{MultiRC}$} & 21.62        & 55.72        \\ \hline
\textbf{Joint Model}          & 20.36        & 57.08        \\ \hline
\textbf{Human}                 & 56.56        & 83.84        \\ \hline
\end{tabular}
\caption{Summary of the results on MultiRC dataset. $EM0$ is the percentage of questions for which all the choices are correct. $EM1$ is the the percentage of questions for which at most one choice is wrong.}
\label{tab:multirc_results}
\end{table}

\indent Results of SemEval experiments are summarized in table \ref{tab:semeval_results}. TriAN-single results are as reported in \cite{yunfudao}. The results we obtained using their implementation are stand-alone QA results. With the same setting, joint model got $85.4\%$ on dev set and $82.1\%$ on test set. The difference in performance of the models in tables \ref{tab:multirc_results} and \ref{tab:semeval_results} is statistically significant according to McNemar's chi-squared test.
\begin{table}[!h]
\centering
\begin{tabular}{|l|l|l|}
\hline
\textbf{Model}                                                                                   & \textbf{Dev} & \textbf{Test} \\ \hline
\textbf{\begin{tabular}[c]{@{}l@{}}TriAN-single\\ \cite{yunfudao} \end{tabular}} & 83.84\%               & 81.94\%                \\ \hline
\textbf{Stand-alone QA} & 83.20\%               & 80.80\%                \\ \hline
\textbf{Joint Model}                                                                             & 85.40\%               & 82.10\%                \\ \hline
\end{tabular}
\caption{Accuracy of various models on SemEval'18 task-11 dataset}
\label{tab:semeval_results}
\end{table}

\section{Discussion}\label{sec:disc}
We have shown that capturing the relationship between various answer choices or subsequent questions helps in answering questions better. Our experimental  results, shown in tables \ref{tab:multirc_results} \& \ref{tab:semeval_results}, are only a first step towards leveraging this relationship to help construct better machine reading systems. We suggest two possible extensions to our model,  that would help realize the potential of these relations.
\begin{enumerate}[noitemsep]
    \item{Improving the performance of entailment and contradiction detection.}
    \item{Using the information given in the text to identify the relations between choices better.}
\end{enumerate}
\indent As shown in table \ref{tab:self_tr}, identification of entailment/contradiction is far from perfect. Entailment detection is particularly worse because often the system returns \textit{entailment} when there is a high lexical overlap. Moreover, the presence of a strong negation word (\textit{not}) causes the NLI system to predict \textit{contradiction} even for \textit{entailment} and \textit{neutral} cases. This issue impedes the performance of our model on SemEval'18 dataset as roughly $ 40\%$ of the questions have \textit{yes/no} answers. \citet{nli-stress-test-2018} show that this is a common issue with state-of-the-art NLI detection models.\\
\indent Self-training (table \ref{tab:self_tr}) results suggest that there are other types of relationships present among answer choice pairs that do not come under the strict definitions of \textit{entailment} or \textit{contradiction}. Upon investigating, we found that although some answer hypotheses do not directly have an inference relation between them, they might be related in context of the given text. For example, consider the sentence, `\textit{I snack when I shop}' and the answer choices: $c_1$: `\textit{She went shopping this extended weekend}' and $c_2$: `\textit{She ate a lot of junk food recently}'. Although the sentences don't have an explicit relationship when considered in isolation, the text suggests that $c_1$ might entail $c_2$. Capturing these kinds of relationships could potentially improve MC further. 
\section{Conclusion}\label{sec:conclu}
In this paper we take a first step towards modeling an accumulative knowledge state for machine comprehension, ensuring consistency between the model's answers. We show that by adapting NLI to the MC task using self-training, performance over multiple tasks improves.\\
\indent In the future, we intend to generalize our model to other relationships beyond strict entailment and contradiction relations.
\section*{Acknowledgements}\label{sec:ack}
\indent We would like to thank the reviewers for their insightful comments.  This work was partially supported by the NSF through grant NSF-1814105.

\end{document}